\documentclass[10pt,twocolumn,letterpaper]{article}

\usepackage{cvpr}              

\usepackage{graphicx}
\usepackage{amsmath}
\usepackage{amssymb}
\usepackage{booktabs}
\usepackage{bm}
\usepackage{multirow}
\usepackage{stfloats}
\usepackage{bbding}
\usepackage{comment}

\usepackage[pagebackref,breaklinks,colorlinks]{hyperref}

\usepackage[capitalize]{cleveref}
\crefname{section}{Sec.}{Secs.}
\Crefname{section}{Section}{Sections}
\Crefname{table}{Table}{Tables}
\crefname{table}{Tab.}{Tabs.}

\def\cvprPaperID{1860} 
\def\confName{CVPR}
\def\confYear{2023}

\begin{document}

\newcommand{\hao}[1]{{\color{blue}[Hao: #1]}}
\newcommand{\jiyao}[1]{{\color{blue}[Jiyao: #1]}}
\newcommand{\yang}[1]{{\color{blue}[Yang: #1]}}
\newcommand{\zekai}[1]{{\color{blue}[Zekai: #1]}}
\newcommand{\verify}[1]{{\color{red}[Verify: #1]}}
\newcommand{\todo}[1]{{\color{red}{\textbf{TODO}: #1}}}

\title{Robot Structure Prior Guided Temporal Attention for Camera-to-Robot Pose Estimation from Image Sequence}

\author{
Yang Tian\textsuperscript{* \rm 1,3},
Jiyao Zhang\textsuperscript{* \rm 1,2,3},
Zekai Yin\textsuperscript{* \rm 1},
Hao Dong\textsuperscript{$\dagger$ \rm 1,2,3} \\
{$^1$ Center on Frontiers of Computing Studies, School of Computer Science, Peking University} \\
{$^2$ Beijing Academy of Artificial Intelligence} \\
{$^3$ National Key Laboratory for Multimedia Information Processing, School of CS, PKU} \\
\small{
\texttt{2301110749@pku.edu.cn},
\texttt{jiyaozhang@stu.pku.edu.cn}, 
\texttt{\{1900017763, hao.dong\}@pku.edu.cn}} \\
}
\maketitle

\begin{abstract}
In this work, we tackle the problem of online camera-to-robot pose estimation from single-view successive frames of an image sequence, a crucial task for robots to interact with the world. 
The primary obstacles of this task are the robot’s self-occlusions and the ambiguity of single-view images.
This work demonstrates, for the first time, the effectiveness of temporal information and the robot structure prior in addressing these challenges.
Given the successive frames and the robot joint configuration, our method learns to accurately regress the 2D coordinates of the predefined robot’s keypoints (\textit{e.g.} joints).
With the camera intrinsic and robotic joints status known, we get the camera-to-robot pose using a Perspective-n-point (PnP) solver.
We further improve the camera-to-robot pose iteratively using the robot structure prior.
To train the whole pipeline, we build a large-scale synthetic dataset generated with domain randomisation to bridge the sim-to-real gap.
The extensive experiments on synthetic and real-world datasets and the downstream robotic grasping task demonstrate that our method achieves new state-of-the-art performances and outperforms traditional hand-eye calibration algorithms in real-time (36 FPS). Code and data are available at the project page: \href{https://sites.google.com/view/sgtapose}{https://sites.google.com/view/sgtapose}.
\end{abstract}

\renewcommand{\thefootnote}{}
\footnote{*: Equal contributions. $\dagger$: Corresponding author}
    \vspace{-0.5cm}
\section{Introduction}
\label{sec:intro}

\begin{figure}[tbp]
\centering
\includegraphics[trim=0 0 0 0,clip, width=\linewidth]{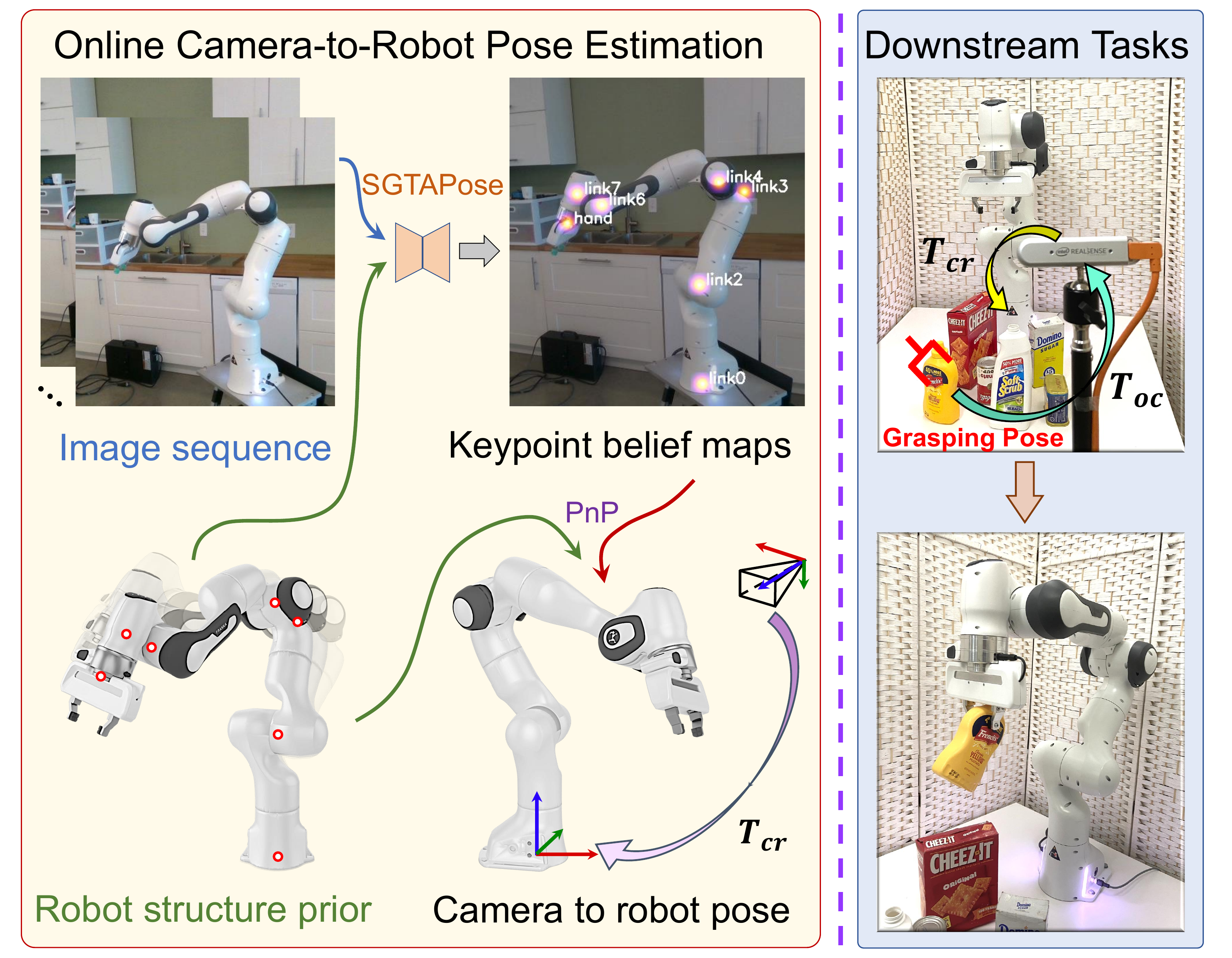}
\caption{\textbf{Overview of the proposed SGTAPose.} 
Given a temporal sequence of RGB frames and known robot structure priors, our method estimates the 2D keypoints (\textit{e.g.}, joints) of the robot and performs real-time estimation of the camera-to-robot pose by combining a \emph{Perspective-n-point} (PnP) solver(left). 
This real-time camera-to-robot pose estimation approach can be utilised for various downstream tasks, such as robotic grasping(right).}
\label{teaser}
\vspace{-6mm}
\end{figure}
Camera-to-robot pose estimation is a crucial task in determining the rigid transformation between the camera space and robot base space in terms of rotation and translation. Accurate estimation of this transformation enables robots to perform downstream tasks autonomously, such as grasping, manipulation, and interaction. Classic camera-to-robot estimation approaches, \textit{e.g.}\cite{fassi2005hand, horaud1995hand, strobl2006optimal}, typically involve attaching augmented reality (AR) tags as markers to the end-effector and directly solving a homogeneous matrix equation to calculate the transformation. However, these approaches have critical drawbacks. Capturing multiple joint configurations and corresponding images is always troublesome, and these methods cannot be used online. These flaws become greatly amplified when downstream tasks require frequent camera position adjustment.

To mitigate this limitation of classic offline hand-eye calibration, some recent works \cite{labbe2021single, lambrecht2021optimizing} introduce vision-based methods to estimate the camera-to-robot pose from a single image, opening the possibility of online hand-eye calibration. Such approaches significantly grant mobile and itinerant autonomous systems the ability to interact with other robots using only visual information in unstructured environments, especially in collaborative robotics \cite{Lee2020CameratoRobotPE}.

Most existing learning-based camera-to-robot pose estimation works\cite{Lee2020CameratoRobotPE, lu2022pose, labbe2021single, Simoni2022SemiPerspectiveDH} focus on single-frame estimation. However, due to the ambiguity of the single-view image, these methods do not perform well when the robotic arm is self-occluded.
Since the camera-to-robot pose is likely invariant during a video sequence and the keypoints are moving continually, one way to tackle this problem is to introduce temporal information. 
However, a crucial technical challenge of estimating camera-to-robot pose temporally is how to fuse temporal information efficiently.
To this end, as shown in Fig.~\ref{teaser}, we propose Structure Prior Guided Temporal Attention for Camera-to-Robot Pose estimation (SGTAPose) from successive frames of an image sequence.
First, we proposed robot structure priors guided feature alignment approach to align the temporal features in two successive frames.
Moreover, we apply a multi-head-cross-attention module to enhance the fusion of features in sequential images.
Then, after a decoder layer, we solve an initial camera-to-robot pose from the 2D projections of detected keypoints and their 3D positions via a PnP solver. 
We lastly reuse the structure priors as an explicit constraint to acquire a refined camera-to-robot pose.

By harnessing the temporal information and the robot structure priors, our proposed method gains significant performance improvement in the accuracy of camera-to-robot pose estimation and is more robust to robot self-occlusion.
We have surpassed previous online camera calibration approaches in synthetic and real-world datasets and show a strong dominance in minimising calibration error compared with traditional hand-eye calibration, where our method could reach the level of 5mm calibration errors via multi-frame PnP solving.
Finally, to test our method's capability in real-world experiments, we directly apply our predicted pose to help implement grasping tasks. We have achieved a fast prediction speed (36FPS) and a high grasping success rate.
Our contributions are summarised as follows:
\begin{itemize}
\vspace{-0.2cm}
\item For the first time, we demonstrate the remarkable performance of camera-to-robot pose estimation from successive frames of a single-view image sequence.
\vspace{-0.2cm}
\item We propose a temporal cross-attention strategy absorbing robot structure priors to efficiently fuse successive frames' features to estimate camera-to-robot pose.
\vspace{-0.2cm}
\item We demonstrate our method's capability of implementing downstream online grasping tasks in the real world with high accuracy and stability, even beyond the performance of classical hand-eye calibration.
\end{itemize}
\vspace{-2mm}
\section{Related Works}
\label{sec:related}

\noindent\textbf{Instance-level 6D Object Pose Estimation.}
Given RGB images and the robot's CAD model,  the camera-to-robot pose is solely determined by the 6D pose of the robot base.
Therefore, our objective is highly correlated with the instance-level 6D rigid object pose estimation \cite{Zhu2022ARO, Kehl2017SSD6DMR}.
Its goal is to infer an object's 6D pose given a reference frame by assuming the exact 3D CAD model is available.
Traditional methods, including iterative closest point (ICP) \cite{besl1992method}, perform template matching by aligning CAD models with the observed pointclouds. 
Some recent works \cite{Song2020HybridPose6O, Lin2021SparseSC, Iwase2021RePOSEF6, 6d1} regard the pose estimation as a regression or a classification task.
2D parameter representation or geometry-guided features will be predicted \cite{Li2019DeepIMDI,Simonyan2015VeryDC,Dosovitskiy2015FlowNetLO}, and then improved PnP solvers \cite{Wang2021PnPDETRTE, Chen2022EProPnPGE} are used to estimate poses.
Although these works are closely related to ours, the manipulator is an articulated object with several degrees of freedom and potential entangling parts, whose pose is tougher to estimate. 

\noindent\textbf{2D Center-based Object Detection and Tracking.}
Our approach predicts camera-to-robot pose via keypoint estimation, which shares similar goals in 2D center-based object detection and tracking.
The center-based method \cite{CenterNet, CenterKT} has been an emerging anchor-free object detection method in recent years,
which models an object as a single point via keypoint estimation and regresses other object properties such as bounding boxes, 3D locations, or poses \cite{ct1, ct2, ct3}. 
Some works \cite{CenterTrack, DEFT, Track, ct4} also extend these center-based models to tracking tasks such as multi-category tracking and pose tracking, applying a detection model to a pair of images and detections from the previous frame.
These center-based methods \cite{ct5, ct6, ct7} have succeeded in simplicity and speed. 
However, the methods mentioned \cite{CenterNet, CenterTrack, CenterPoint} mainly adopt simple concatenations to fuse temporal information and thus overlook accurate pixel-wise correspondence, where we propose a robot structure guided feature alignment module and a temporal cross-attention module to produce better feature fusion.

\noindent\textbf{Robot Arm Pose Estimation.}
Recently, many learning-based camera-to-robot pose estimation methods have been proposed, which can be divided into three types.
The first type falls into keypoint-based methods via a single RGB image.
For example, DREAM\cite{Lee2020CameratoRobotPE} designs a CNN-based pipeline to regress 2D keypoints, construct 2D-3D correspondence and recover the camera-to-robot pose via a PnP-solver. 
Lately, \cite{lu2022pose} seeks to find the optimal 2D keypoint candidates for better acquiring pose estimations.
The second type falls into rendering-based methods given robots' 3D CAD models \cite{rb1, labbe2021single}. 
Robopose \cite{labbe2021single} optimises the camera-to-robot pose by iteratively rendering images and comparing them with ground truth.
This method requires a long time (1s) to predict an excellent initial pose and fails in dynamic scenarios. 
The last type falls into depth-based methods \cite{rb2,Simoni2022SemiPerspectiveDH}, which rely highly on depth sensors' precision and lack high accuracy in real-world experiments.

\section{Method}
\label{sec:method}

\begin{figure*}[tbp]
\centering
\vspace{-15pt}

\includegraphics[trim=0 0 0 0,clip, width=\linewidth]{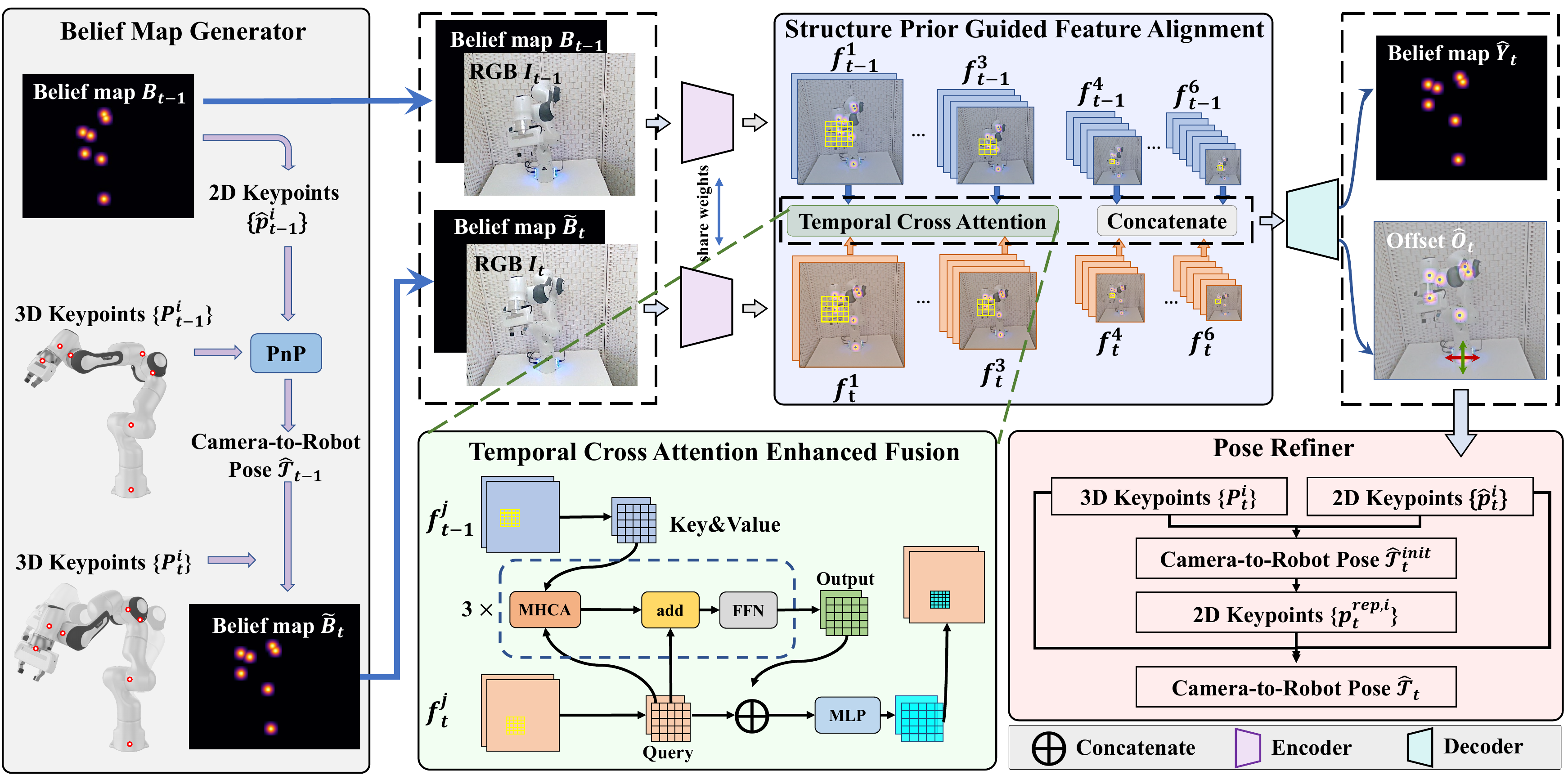}
\caption{\textbf{Pipeline of the proposed SGTAPose.} (a) \textbf{Structure Prior Guided Feature Alignment} We use a \textbf{Belief Map Generator} to acquire belief maps as the structure prior.
Given 2D/3D keypoints' locations $\{\hat{p}_{t-1}^i\}$ and $\{P_{t-1}^i\}$ from the previous frame $I_{t-1}$, an initial camera-to-robot pose is calculated and used to project current keypoints' 3D locations $\{P_t^i\}$ to a reprojected belief map $\Tilde{B}_t$. The paired inputs are sent to a shared encoder and yield multi-scale features $f_{t-1}^j$ and $f_{t}^j, j \in [6]$. (b) \textbf{Temporal Cross Attention Enhanced Fusion} For $j\in \{1,2,3\}$, we implement temporal cross attention in the regions around the detected keypoints in $B_{t-1}$ and $\Tilde{B}_t$ to efficiently fuse features from successive frames. 
(c) \textbf{Pose Refiner.} We acquire the detected keypoints' 2D locations and solve an initial camera-to-robot pose $\hat{\mathcal{T}}_{t}^{init}$. Then we gain a refined camera-to-robot pose $\hat{\mathcal{T}}_{t}$ with structure priors via a weighted Levenberg-Marquardt (LM) solver.}
\label{pipeline}
\vspace{-3mm}
\end{figure*}


We will first introduce the problem statement in Section \ref{Problem Statement}.
Then we will explain three modules, \textbf{Structure 
 Prior Guided Feature Alignment}, \textbf{Temporal Cross Attention Enhanced Fusion}, and \textbf{Pose Refiner} in our approach.
\textbf{Structure Prior Guided Feature Alignment} aims to align corresponding features between frames and is discussed in Section \ref{Robot Structure Prior}. \textbf{Temporal Cross Attention Enhanced Fusion} targets fusing temporal information and is discussed in Section \ref{Temporal CA}. Finally, we describe \textbf{Pose Refiner} in Section \ref{C2RP Refinement} to refine the predicted pose better. 

\subsection{Problem Statement} \label{Problem Statement}
Our problem is defined as follows: Given a live stream of RGB images $\{I_t\}_{t \geq 0}$ containing a manipulator along with its instant camera-to-robot pose $\{\mathcal{T}_t\}_{t \geq 0} = \{ (R_t, T_t) \}_{t \geq 0}$ $\in SE(3)$, the transformation from the camera space to the robot base space,  
our objective is to track the camera-to-robot pose in an online manner. 
In other words, at timestep $t$, provided images $I_{t-1}$ and $I_{t}$, predefined keypoints' 3D positions $\{P_t^{i}\}$ in robot space ($i$ denotes the index of each keypoint), camera intrinsics
and the estimated pose $\hat{\mathcal{T}}_{t-1}$, we predict the rotation matrix $R_t$ and translation $T_t$ in $\hat{\mathcal{T}}_{t}$.

\subsection{Structure Prior Guided Feature Alignment}\label{Robot Structure Prior}
Inspired by the fact that the joints' states and pose of the robot change slightly between successive two frames, we believe the estimated previous pose $\hat{\mathcal{T}}_{t-1}$ will work as solid structure priors for guiding network learning. 
In this way, our first step is to design a \textbf{Belief Map Generator} to produce a reprojection belief map. 
Suppose we have acquired the previous estimated pose $\hat{\mathcal{T}}_{t-1}$, we reproject the instant keypoints' 3D positions $\{P_t^{i}\}$ into $\{\Tilde{p}_{t}^{i}\}$, and visualise them into a single-channel belief map $\Tilde{B}_{t} \in [0, 1]^{H \times W \times 1}$ (See in the left column of Figure \ref{pipeline}) . 
Based on $\Tilde{B}_{t}$, the network cares more about the residuals of $\Tilde{B}_{t}$ to its ground truth.

Similarly, we also prepare the previous belief map $B_{t-1}$ for guiding the network to align the previous frame's features.
To be more specific, during the training process, $B_{t-1}$ is produced by augmenting ground truth 2D keypoints.
While during inference, we utilise 2D keypoints $\{\hat{p}_{t-1}^{i}\}$ estimated from previous network output belief map $\hat{Y}_{t-1} \in \mathbb{R}^{\frac{H}{R} \times \frac{W}{R} \times c}$ and $\hat{O}_{t-1} \in \mathbb{R}^{\frac{H}{R} \times \frac{W}{R} \times 2}$ where $R$ and $c$ denotes the downsampling ratio and amounts of keypoints.


Then, we send the pairs $(I_{t-1}, B_{t-1})$ and $(I_t, \Tilde{B}_t)$ to a shared backbone and obtain two lists composed of 6 multi-scale features $L_{t-1}$ and $L_{t}$. We denote $L$ (for simplicity, we omit the subscript t) as $[f^1, \dots, f^6]$ where $f^j \in \mathbb{R}^{c_j \times h_j \times w_j}$ and $\frac{c_{j+1}}{c_j} = \frac{h_j}{h_{j+1}} = \frac{w_j}{w_{j+1}} = 2$. 
$\{\hat{p}_{t-1}^{i}\}$ and $\{\Tilde{p}_{t}^{i}\}$ are rescaled to match the $f$'s size accordingly and treated as center proposals. 
Features extracted from the neighbourhoods of center proposals in $f_{t-1}$ and $f_t$ are regarded as aligned since they roughly entail the same keypoint's contextual information.
The aligned features should be fused carefully, illustrated in the next section.

\subsection{Temporal Cross Attention Enhanced Fusion}\label{Temporal CA}
To carefully integrate the multi-scale aligned features in $L_{t-1}$ and $L_t$, we adopt different strategies considering the size of $f^j$ 
For $f^m, m \in \{1,2,3\}$, they have much higher resolutions and fine-grained features, which are crucial for detecting small-sized keypoints.
Therefore, we propose a temporal cross-attention module to fuse the features at the neighbourhood of center proposals.
In comparison, for $f^n, n \in \{4,5,6\}$, they have lower resolutions and a broader receptive field, so each pixel-level feature contains more contextual and temporal information.
We thus directly concatenate the features at the center proposals of $f_{t-1}^n$ and $f_{t}^n$ and process them into original sizes $\mathbb{R}^{c \times c_n}$ via a shallow Multilayer Perceptron (MLP). The newly processed features will instantly replace the counterpart in $f_{t}^n$.

For the first three feature maps $\{f^m, m \in \{1,2,3\} \}$, 
we treat $\{\hat{p}_{t-1}^{i}\}$ and $\{\Tilde{p}_{t}^{i}\}$ as center proposals respectively, and rescale them to match the size of $f^m$. 
Having measured the motion amplitude of the manipulator in finishing downstream tasks (e.g., Grasping), we confine the center proposals to a square area with window size $d_m$.
We take $f_{t-1}^m$ at the $d_m \times d_m$ window area around scaled $\{\hat{p}_{t-1}^{i}\}$ as query embeddings and $f_{t}^m$ at the same size of area around $\{\Tilde{p}_{t}^{i}\}$ as keys and values.
After 3 vanilla Tranformer multi-head cross-attention layers  \cite{vaswani2017attention}, we concatenate the output features $Q_{t, d_m^2c}^m \in \mathbb{R}^{d_m^2c \times c_m}$ with $f_{t, d_m^2c}^m$ at the same $d_m^2c$ locations along the $c_m$-dimension, and send them through a shallow MLP to get $\hat{f}_{t, d_m^2c}^m$. 
We directly replace  $f_{t, d_m^2c}^m$ with $\hat{f}_{t, d_m^2c}^m$ and pass all the six processed multi-scale features to the decoder layer and receive the output head.

\subsection{Pose Refiner}\label{C2RP Refinement}
We design a pose refiner to mitigate the influence of outlier keypoints with significant reprojection errors when computing the camera-to-robot pose.
Since the initial pose solved by \emph{Perspective-n-Point} (PnP) algorithms might be inaccurate sometimes due to outliers \cite{Ferraz2014VeryFS}, we correct such bias by solving reweighted PnP problems.
Utilising predicted projections $\{\hat{p}_{t}^i\}$ and known $\{P_{t}^{i}\}$, we obtain an initial camera-to-robot pose $\hat{\mathcal{T}}_{t}^{init}$ via a PnP-RANSAC solver \cite{PNPRANSAC}.
Next, we project $\{P_t^i\}$ via $\hat{\mathcal{T}}_{t}^{init}$ to 2D coordinates $\{p_t^{rep, i}\}$.
We set the weights $\omega_t^i = \exp{(-5 \times \Vert \hat{p}_{t}^i -  p_t^{rep, i} \Vert^2 )}$ based on practical experience and optimise the following equation via an LM solver \cite{Pesaran2008ABL}.
\begin{equation}
\begin{aligned}
\arg \min_{R_t,  T_t} \frac{1}{2}\sum_{i=1}^{c} \Vert \omega_t^i (\pi (R_t P_t^{i} + T_t ) - \hat{p}_t^{i}) \Vert ^2 \label{3D Rf}
\end{aligned}
\end{equation}
where $\pi(\cdot)$ is the projection function, $R_t$ and $T_t$ is the rotation and translation in the camera-to-robot pose $\hat{\mathcal{T}}_{t}$ .The reweighted optimisation objective focuses more on the "influence" of comparatively precise predictions, thus mitigating the impact of keypoints with large reprojection errors. 

\subsection{Implementation details} \label{Id}
\noindent\textbf{Loss function}. 
Our network output involves a predicted pixel-level belief map $\{\hat{Y}_t\} \in [0,1]^{\frac{H}{R} \times \frac{W}{R} \times c}$ and subpixel-level local offsets $\{\hat{O}_t\} \in [0,1]^{\frac{H}{R} \times \frac{W}{R} \times 2}$.
We design two loss functions $L_B$ and $L_{off}$ for $\{\hat{Y}_t\}$ and $\{\hat{O}_t\}$ respectively. 
For keypoints' ground truth 2D locations $\{p_t^i\}$, we scale them into a low-resolution equivalence $p_{low, t}^i= \lfloor \frac{P_t^i}{R} \rfloor$. We draw each keypoint in a single-channel feature map with a Gaussian Kernel $K(x, y) = \exp{(-\frac{(x - p_{low, t,x}^i)^2 + (y - p_{low, t, y}^i)^2}{8})}$ and shape the ground truth belief map $Y_t \in [0,1]^{\frac{W}{R} \times \frac{H}{R} \times c}$.
The $L_B$ becomes: 
\begin{equation} 
    \begin{aligned}
        L_B =  \Vert Y_t - \hat{Y}_t \Vert_{L_2}^2
    \end{aligned}
\end{equation}
\label{L_B}
where $R = 4$ in our network.
We further follow \cite{CenterTrack} to correct the error induced by the output stride. The offsets$\{\hat{O}_t\}$ are trained via smooth $L_1$ loss and only supervised in locations $p_{low}^i$:
\begin{equation}
    \begin{aligned}
        L_{off} = \Vert \hat{O}_{t}{}_{p_{low, t}^i} - (\frac{p_t^i}{R} - p_{low, t}^i) \Vert
    \end{aligned}
\end{equation}
\label{L_off}
The overall training objective is designed as follows:
\begin{equation}
    \begin{aligned}
        L = \lambda_B L_B + \lambda_{off} L_{off}
    \end{aligned}
\end{equation}
\label{Loss Function Equation}
where $\lambda_B$ = 1.0 and $\lambda_{off} = 0.01$ in implementation.

\vspace{3mm}
\noindent\textbf{Training details.}
During the training time, we pre-process the input image $I_{t-1}, I_t$ into the size of $\mathbb{R}^{480 \times 480 \times 3}$ via affine transformation and normalisation with mean $[0.5, 0.5, 0.5]$ and standard deviation $[0.5, 0.5, 0.5]$.
To further improve our model's robustness, we apply $\mathcal{N}(0, 1.5I)$ noises as well as randomly drop with probability 0.2 to the ground truth keypoints in $B_{t-1}$.
Our backbone is based on Deep Layer Aggregation \cite{Yu2018DeepLA} and trained for 20 epochs with batch size 16, Adam optimiser \cite{Kingma2015AdamAM} with momentum 0.9 and 180k synthetic training images.
The learning rate warms up to 1.25e-4 from 0 during the first 3,000 iterations and drops to 0 during the rest of the iterations linearly.

\vspace{2mm}
\noindent\textbf{Inference details.}
During inference, we are given a long-horizon video split into consecutive frames.
We use the first frame as $I_0$ and $I_1$, and blank images as the initial belief maps $B_0$ and $B_1$ to perform inference. 
For each timestep $t > 1$, we select the keypoints' 2D locations $p_{low}^i$ with the largest confidence score  for each belief map in $\hat{Y}_{t-1} \in \mathbb{R}^{\frac{H}{R} \times \frac{W}{R} \times c}$.
We then determine the accurate locations by adding $\hat{O}_{t}{}_{p_{low, t}^i}$ to $p_{low, t}^i$. 
Finally, we rescale these low-resolution keypoints' locations to match the raw image's size via inverse affine transformation and obtain $\{\hat{p}_t^i\}$.

\section{Experiments}
\label{sec:experiments}

\subsection{Datasets}

\begin{figure}
\centering
\includegraphics[trim=0 0 0 0,clip, width=0.95\linewidth]{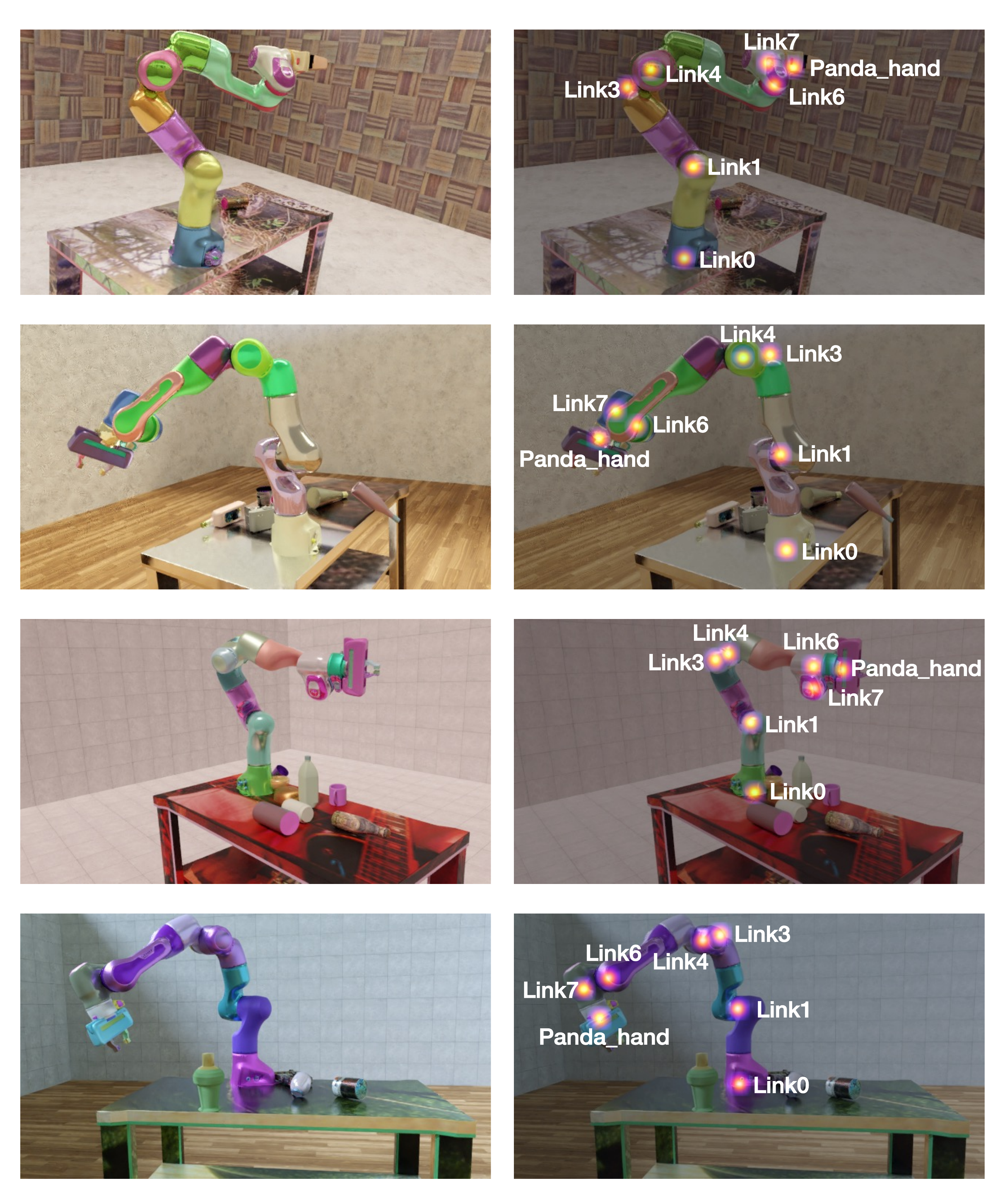}
\vspace{-3mm}
\caption{Sample images from Panda Syn Training (first column) and annotations (second column). We set every part a random colourful tint to improve the diversity of our datasets and apply several domain randomisations to shorten the sim-to-real gap.}
\label{Dataset}
\vspace{-6mm}
\end{figure}

Our datasets involve one self-generated synthetic training set (Panda Syn Training), one self-generated synthetic testing set (Panda Syn Testing), and three real-world testing sets (Panda 3CAM-RS, Panda 3CAM-AK, and Panda Orb) provided by \cite{Lee2020CameratoRobotPE}.  
Since the training set proposed in \cite{Lee2020CameratoRobotPE} doesn't support temporal images, we thus generate Panda Syn Training/Testing.
The three public real-world sets are generated by an externally mounted camera filming a Franka Emika Panda manipulator according to \cite{Lee2020CameratoRobotPE}.
Panda 3CAM-AK is collected by a Microsoft Azure Kinect camera, and Panda 3CAM-RS/Orb are collected by Intel RealSense D415.
Panda 3CAM-AK and Panda 3CAM-RS are captured from a single fixed view and involve about 6k images respectively, while Panda Orb is captured from 27 different views and involves approximately 32k images.

Following~\cite{dai2022domain}, we construct large-scale synthetic sets Panda Syn Training/Testing (Figure \ref{Dataset}) utilising Blender~\cite{Blender}. 
We generate frames in a video at FPS 30 with a fixed scene and a moving manipulator. 
Several domain randomisations have been used to shorten the sim-to-real gap and will be detailedly explained in supplementary materials.
The synthetic dataset consists of temporal RGB images, 2D/3D predefined keypoints' locations, and part poses.
Overall, Panda Syn Training involves approximately 60k videos which contain 3 successive frames per video (180k images), and Panda Syn Testing involves 347 videos which contain 30 successive frames per video (10k images).

\subsection{Baselines and Evaluations}
\noindent\textbf{Baselines.} 
We compare our framework with previous online keypoint-based camera-to-robot pose estimation approaches and center-based object detection methods via single or multi frames. Besides, we also compare our framework with traditional offline hand-eye calibration.
\textbf{Dream}\cite{Lee2020CameratoRobotPE}: A pioneering approach for estimating the camera-to-robot pose from a single frame by direct 2D keypoints regression and PnP-RANSAC solving.
\textbf{CenterNet}\cite{CenterNet}: A single-frame object detection approach that models an object as a single point. We adapt CenterNet to estimate 2D keypoints and reserve the heatmap loss and offset loss \cite{CenterNet} during training. 
\textbf{CenterTrack}\cite{CenterTrack}: A multi-frame object detection and tracking approach. We adapt CenterTrack using similar strategies in CenterNet to detect keypoints.

\noindent\textbf{Metrics.} During the inference, we evaluate both 2D and 3D metrics across all datasets.
\textbf{PCK}: The L2 errors between the 2D projections of predicted keypoints and ground truth. 
Only keypoints that exist within the frame will be considered.
\textbf{ADD}: The average Euclidean norm between 3D locations of keypoints and corresponding transformed versions, which directly measures the pose estimation accuracy.
For PCK and ADD, we compute the area under the curve (AUC) lower than a fixed threshold (12 pixels and 6cm, respectively) and their median values.


\subsection{Results and Analysis}

As can be seen in Table \ref{PCK}, our method has outperformed all other baselines in PCK and ADD metrics across datasets.
Specifically, 
our median PCK, an important measure that reflects the 2D accuracy, is superior to others, indicating that we have achieved better overall precision in predicting the 2D projections.
We also surpass all other baselines in the AUC of ADD, and so is the median. ADD is a more direct indicator of whether the estimated pose is accurate, and we have outperformed the best of others in Panda 3CAM-AK and Panda Orb at 4.9\% and 7.8\%.
Further, our method can reach 9.77mm and 18.12mm in median ADD in Panda 3CAM-RS and Panda Orb, comparable to the accuracy of the traditional hand-eye calibration empirically.

Compared with single-frame methods (CenterNet\cite{CenterNet}, Dream \cite{Lee2020CameratoRobotPE}), our method absorbs temporal information and shows greater robustness to self-occlusion. The analysis of robustness to self-occlusion is discussed in Section \ref{Robustness to Self Occlusion}. 
Meanwhile, compared with the tracking-based method CenterTrack \cite{CenterTrack}, our method owns the robot structure prior guiding feature alignment and temporal cross attention, which we believe facilitates our model's superiority. 

\begin{table*}[]
\centering
\resizebox{\linewidth}{!}{
\begin{tabular}{ccccccccc}\toprule
\multirow{2}{*}{Dataset} & \multirow{2}{*}{Real Data}  &  \multirow{2}{*}{\# Images} & \multirow{2}{*}{\# 6D Poses} & \multirow{2}{*}{Method} & \multicolumn{2}{c}{PCK} & \multicolumn{2}{c}{ADD}  \\ 
\cline{6-9}
    ~ &   ~ &    ~ &  ~ &        ~ & AUC$\uparrow$ ~ & Median@pix$\downarrow$~ & AUC$\uparrow$ ~ & Median@mm$\downarrow$ \\ 
\hline
\multirow{4}{*}{Panda 3CAM-RS\cite{Lee2020CameratoRobotPE}} & \multirow{4}{*}{\checkmark} & \multirow{4}{*}{5944} & \multirow{4}{*}{1}
            & CenterNet\cite{CenterNet}             & 67.38         & 3.51         & 59.26         & 21.25         \\ 
    ~ &   ~ &  ~ &  ~ & CenterTrack\cite{CenterTrack}         & 68.85         & 3.59         & 58.24         & 23.77  \\ 
    ~ &   ~ &  ~ &  ~ &  Dream\cite{Lee2020CameratoRobotPE}    & 64.82         & 3.90         & 58.60         & 24.57  \\ 
    ~ &   ~ &  ~ &  ~ & Ours                                  &\textbf{75.68} &\textbf{2.68} &\textbf{79.89} &\textbf{9.77} \\ \hline
\multirow{4}{*}{Panda 3CAM-AK\cite{Lee2020CameratoRobotPE}} & \multirow{4}{*}{\checkmark} & \multirow{4}{*}{6394}  & \multirow{4}{*}{1}
            & CenterNet\cite{CenterNet}             & 52.38         & 4.90         & 34.07         & 37.56 \\ 
    ~ &   ~ &  ~ &  ~ & CenterTrack\cite{CenterTrack}         & 58.26         & 4.45         & 43.10         & 32.83 \\ 
    ~ &   ~ &  ~ &  ~ &  Dream\cite{Lee2020CameratoRobotPE}    & 52.28         & 4.83         & 44.55        & 33.68   \\ 
    ~ &   ~ &  ~ &  ~ & Ours                                  &\textbf{62.75} &\textbf{3.19} &\textbf{49.42} &\textbf{29.61} \\ \hline
\multirow{4}{*}{Panda Orb\cite{Lee2020CameratoRobotPE}}& \multirow{4}{*}{\checkmark}  & \multirow{4}{*}{32315}  & \multirow{4}{*}{27}
            & CenterNet\cite{CenterNet}             & 60.11         & 3.47         & 50.59         & 24.22 \\ 
    ~ &   ~ &  ~ &  ~ & CenterTrack\cite{CenterTrack}         & 61.03         & 3.73         & 47.67         & 25.13  \\ 
    ~ &   ~ &  ~ &  ~ & Dream\cite{Lee2020CameratoRobotPE}    & 57.44        & 3.73         & 52.56         & 22.53   \\ 
    ~ &   ~ &  ~ &  ~ & Ours        &\textbf{63.28} &\textbf{3.46} &\textbf{60.30} &\textbf{18.12} \\ \hline
\multirow{4}{*}{Panda Syn Testing} & \multirow{4}{*}{$\times$}  & \multirow{4}{*}{10410}  & \multirow{4}{*}{347}      
            & CenterNet\cite{CenterNet}             & 92.60         & 0.89         & 85.97         & 5.79  \\ 
    ~ &   ~ &  ~ &  ~ & CenterTrack\cite{CenterTrack}         & 91.86         & 0.63         & 85.01         & 6.18  \\ 
    ~ &   ~ &  ~ &  ~ & Dream\cite{Lee2020CameratoRobotPE}    & 80.79         & 0.85         & 79.05         & 7.13  \\ 
    ~ &   ~ &  ~ &  ~ & Ours        &\textbf{94.36} &\textbf{0.44} &\textbf{89.62} &\textbf{4.11} \\
\bottomrule
\end{tabular}}
\caption{\textbf{Quantitative comparison with keypoint-based methods.} $\uparrow$ means higher is better, $\downarrow$ means lower is better. The $\checkmark$ and $\times$ in Real Data denote whether the dataset is real-world. \# Images and \# 6D Poses  denote the total amounts of images and 6D poses in the dataset respectively. For a fair comparison, we train all the methods listed in the above table on Panda Syn Training dataset and report the results regarding PCK and ADD across four testing datasets. Obviously, our method is taking the lead in all metrics upon all datasets.}
 \vspace{-4mm}
\label{PCK}
\end{table*}

\subsection{Compare with Classic Hand-Eye Calibration}
\begin{figure}[h]
\centering
\includegraphics[trim=0 0 0 0,clip, width=0.92\linewidth]{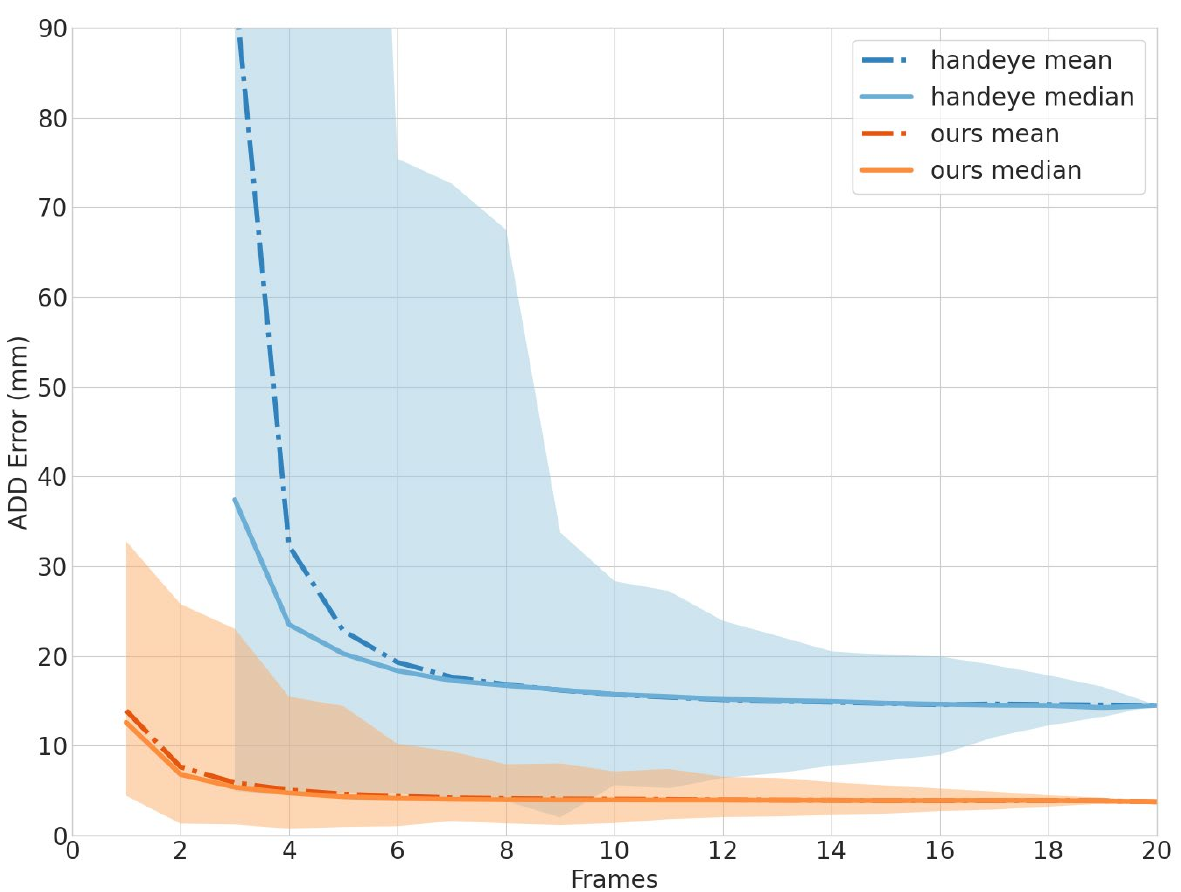}
\caption{Comparison between classic hand-eye calibration and our approach. Results regarding ADD are calculated with increasing amounts of used frames. Mean ADD (dashed lines), median ADD (solid lines), and values between minimum and maximum (shaded areas) have been demonstrated.
}
\label{Hand_eye_ours}
\end{figure}

We further design an experiment to compare our approach with traditional hand-eye calibration (HEC) methods, implemented via the easy\_handeye ROS package \cite{EasyHandEye}.
We attach an Aruco Fiducial Marker \cite{Marker} to the Franka Emika Panda's gripper and place an external RealSense D415 to take photos.
The manipulator is commanded to move to $L=20$ positions along a predefined trajectory and stays at each position for 1 second.
To assess the accuracy, we consider choosing $l$ positions from the set of $L$ positions and feed all the detection results from the $l$ positions to a single PnP solver.
Specifically, we choose $C_{L}^{l}$ combinations of $l$ images and if $C_{L}^{l} > 2500$, we randomly sample 2500 combinations.
As shown in Fig \ref{Hand_eye_ours}, our approach and traditional HEC solve a more accurate pose with increasing frames. 
However, our approach outperforms HEC in both median and mean ADD, where we can reach the level of 5 mm while HEC reaches 15 mm at last.
Moreover, our approach is more stable than HEC with frames increasing, which can be inferred from the region surrounded by the minimum and maximum ADD.

\subsection{Ablation Study}


We conduct the ablation studies to investigate : 
1) The function of the robot structure prior for guiding feature alignment.
2) The necessity of using a cross attention module for enhancing the fusion of temporal aligned features between successive frames. 
3) The effectiveness of introducing the pose refiner module for updating a more accurate pose.
4) The influence of different window sizes adopted during temporal cross attention. 
To this end, we construct four ablated versions of our model and test their performances on the most diverse real dataset Panda Orb \cite{Lee2020CameratoRobotPE}.

\vspace{-1mm}
\begin{table}[t]
\centering
\resizebox{\linewidth}{!}{
\begin{tabular}{ccccccc} \toprule
 \multirow{2}{*}{SGF} & \multirow{2}{*}{TCA}& \multirow{2}{*}{PRF} & \multicolumn{2}{c}{PCK}     & \multicolumn{2}{c}{ADD}   \\
 \cline{4-7}
         ~ &   ~ &  ~        & AUC$\uparrow$ ~ & Median@pix$\downarrow$ ~ & AUC$\uparrow$ ~ & Median@mm$\downarrow$ \\\hline
 & &  & 49.78 & 4.69 & 37.93 & 34.48 \\
$\checkmark$ &  &  & 62.86 & 3.49 & 55.83 & 21.04 \\
$\checkmark$ & $\checkmark$ &  & \textbf{63.28} & \textbf{3.46} & 58.86 & 19.17 \\
$\checkmark$ & $\checkmark$ & $\checkmark$ & \textbf{63.28} & \textbf{3.46} & \textbf{60.30} & \textbf{18.12} \\
\bottomrule
\end{tabular}}
\caption{Ablation studies with different modules. SGF, TCA, and PRF denote the \textbf{Structure Prior Guided Feature Alignment}, \textbf{Temporal Cross Attention Enhanced Fusion}, and \textbf{Pose Refiner} modules, respectively.
$\checkmark$ denotes the corresponding module that has been used. Results show that all the proposed modules benefit camera-to-robot pose estimation.}
\label{Ablation_all}
\vspace{-3mm}
\end{table}

We report the results in Table \ref{Ablation_all}.
For the version without any module, we send $B_{t-1}, I_{t-1}, I_t$ as input with a shared encoder, directly concatenating the multi-scale features.
With the addition of \textbf{Structure Prior Guided Feature Alignment}, we add one more input $\Tilde{B}_t$, and the performance in both PCK and ADD improves greatly.
This fact reveals that the reprojection belief map $\Tilde{B}_t$ provides a confined area rather than the whole image for the network to focus on, which is easier to detect keypoints.
Secondly, with the addition of \textbf{Temporal Cross Attention Enhanced Fusion}, we replace the direct concatenation with cross attention.
The performance in ADD improves higher than that in PCK, which shows that the temporal cross attention module facilitates the precision of 3D keypoint predictions.
Thirdly, the final introduction of \textbf{Pose Refiner} witnesses an improvement of 1.5$\%$ in the AUC of ADD and 1mm in median ADD.
This proves that the reweighted optimisation objective (Equation \ref{3D Rf}) succeeds in filtering out the disturbance of outliers and recomputing a better solution.

As shown in Table~\ref{Ablation_ws}, we change the window size during \textbf{Temporal Cross Attention}.
We analyse the average movement amplitude of a manipulator between two successive frames at FPS 30 and determine the window sizes of the first three multi-scale features.
We ultimately select the window size $[13,7,3]$ to ensure most of the keypoints' movement between two frames can be detected within the window. Results show that the larger or smaller window size will degrade the model's performance in both AUC of ADD and PCK.
We believe the reason for the degradation is that a larger window size will contain more redundant information, while a smaller window size might miss some important information during fast movement tracking.
\vspace{-2mm}
\begin{table}[h]
\centering
\resizebox{\linewidth}{!}{
\begin{tabular}{ccccc} \toprule
\multirow{2}{*}{Window Size $[d_1, d_2, d_3]$}  & \multicolumn{2}{c}{PCK} & \multicolumn{2}{c}{ADD}  \\
\cline{2-5}
        ~  & AUC$\uparrow$ ~ & Median@pix$\downarrow$ ~ & AUC$\uparrow$ ~   & Median@mm$\downarrow$  \\ \hline
$[7,3,1]$ & 61.29 & \textbf{3.41} & 57.11 & 18.21 \\
$[13,7,3]$ & \textbf{63.28} & 3.46 & \textbf{60.30} & \textbf{18.12} \\
$[17,9,5]$ & 63.01 & 3.53 & 56.54 & 21.45\\
\bottomrule
\end{tabular}}
\caption{Results of our ablation study with different window sizes during \textbf{Temporal Cross Attention}. The $[13,7,3]$ window size performs best among all choices on Panda Orb \cite{Lee2020CameratoRobotPE}.}
\label{Ablation_ws}
\vspace{-5mm}
\end{table}

\subsection{Robustness to Self-Occlusion Scenarios}\label{Robustness to Self Occlusion}
\vspace{-2mm}

\begin{figure*}[ht]
\centering
\includegraphics[trim=0 0 0 0,clip, width=\linewidth]{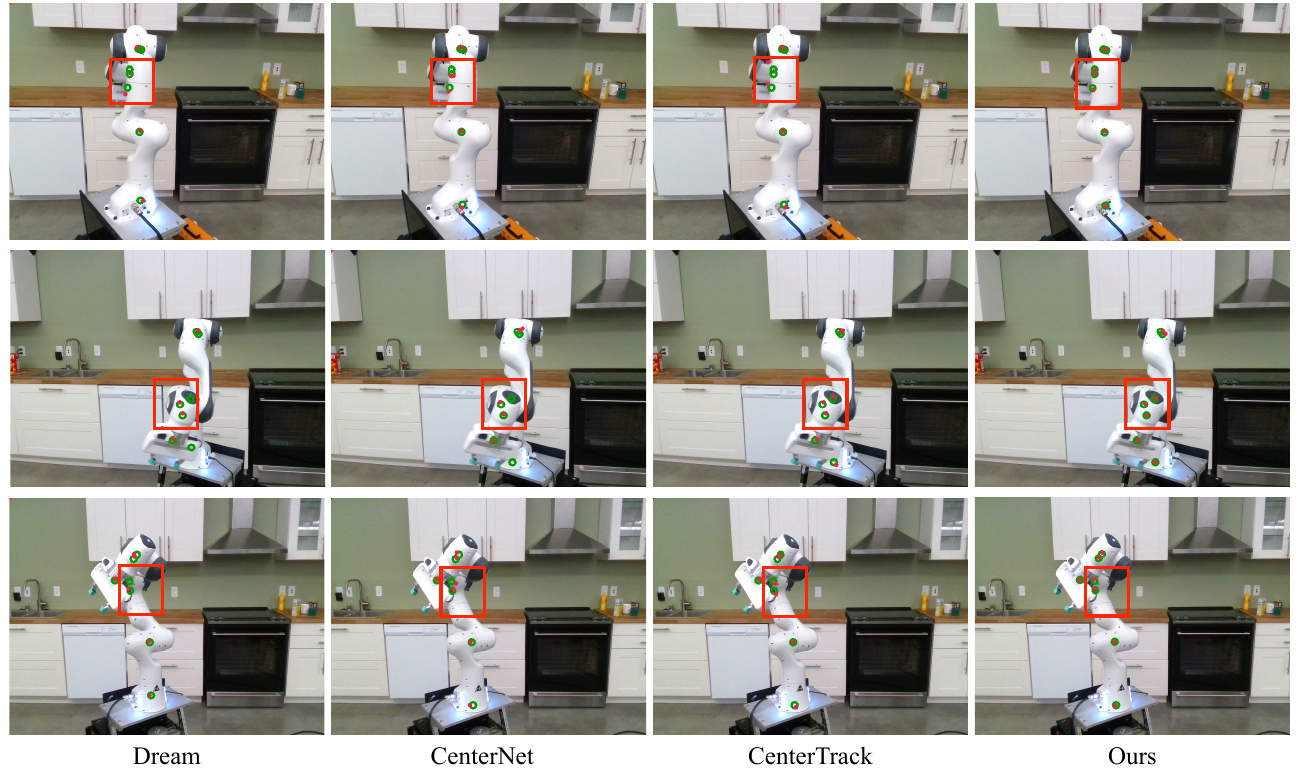}
\caption{Comparison with keypoint-based methods in severe self-occlusion scenarios. For a fair comparison, we retrain all the baselines on Panda Syn Training and show visualisations when encountering severe self-occlusion on the largest dataset Panda Orb \cite{Lee2020CameratoRobotPE}.
The green circles and red points denote the ground truth and estimated keypoints, respectively. Red boxes highlight occluded regions where our method performs much better than all baselines.
This vividly demonstrates that our method is more robust to self-occlusion.}
\label{Self-Occlusion-Figure}
\vspace{-4.5mm}
\end{figure*}

\begin{table}[ht]
\centering
\resizebox{\linewidth}{!}{
\begin{tabular}{ccccc} \toprule
   \multirow{2}{*}{Method} & \multicolumn{2}{c}{PCK} & \multicolumn{2}{c}{ADD}   \\ 
  \cline{2-5}  
  ~ & AUC$\uparrow$ ~ & Median@pix$\downarrow$~ & AUC$\uparrow$ ~ & Median@mm$\downarrow$ \\ \hline
   Dream \cite{Lee2020CameratoRobotPE} & 49.02/59.46 & 4.13/3.67 & 43.87/54.53 & 30.23/21.28  \\
   CenterNet \cite{CenterNet} &  52.86/61.77 & 4.00/3.39 & 39.05/53.21 & 33.73/22.62  \\
   CenterTrack\cite{CenterTrack} & 54.43/62.54  & 4.31/3.62 & 30.37/51.58 & 42.96/22.52  \\
   Ours &  \textbf{60.02}/\textbf{64.03} & \textbf{3.95}/\textbf{3.37} & \textbf{59.37}/\textbf{60.51} & \textbf{20.18}/\textbf{17.69} \\
\bottomrule
\end{tabular}}
\vspace{-3mm}
\caption{Quantitative results of the self-occlusion experiment. The Left and right sides of '/' are the results of severe and no self-occlusion respectively. Results show our method performs robustly when encountering self-occlusion while baselines drop greatly.}
\label{Self-Occlusion-Table}
\vspace{-3mm}
\end{table}
We perform an additional experiment to show our model's robustness to self-occlusion. We divide Panda Orb \cite{Lee2020CameratoRobotPE} into 89 videos, including 31 videos with severe self-occlusion (5971 images) and 58 videos with less or no self-occlusion (26344 images). 
Results in Table \ref{Self-Occlusion-Table} show baselines drop greatly when encountering occlusion, while ours decreases slightly.
Also, from Figure \ref{Self-Occlusion-Figure}, we can see that baselines detect occluded keypoints with significant deviation or even fail. 
In comparison, our model shows more precise predictions. 
The main reason for our model's robustness is that we adopt temporal information fusion and utilise structure priors efficiently rather than estimating pose from a single frame or concatenating temporal features crudely. 

\subsection{Robot Grasping Experiments}
In this section, we construct real-world robotic grasping experiments to demonstrate the performance of our method.

\noindent
\textbf{Experimental Protocol.}
We perform two experiments using the Franka Emika Panda robot. One of the experiments focuses on robot grasping in a static environment, and another on grasping in a dynamic environment. 
To ensure a fair comparison, GraspNet\cite{Fang2020GraspNet1BillionAL} is used to estimate the robot grasping pose in all experiments, while different methods are employed for camera-to-robot pose estimation. 
In the experiments, all learning-based camera-to-robot pose estimation methods use 30 frames to estimate the camera-to-robot pose, and all the objects are selected from YCB~\cite{calli2015benchmarking} dataset.
For hand-eye calibration, in the static experiment, we use easy\_handeye\cite{EasyHandEye} to acquire the pose. In the dynamic experiment, hand-eye calibration is incapable of online calibration, so their results are none.

In the static experiment, we conduct six scenes, and each scene includes  4-7 randomly chosen objects. During the completion of a grasping task in one scene, the camera remains stationary. After finishing grasping in a particular scene, we move the camera to another position for the next scene. 
In the dynamic experiment, we not only change the camera pose when switching between different scenes but also adjust the camera pose after completing the grasping of each object during the execution of a grasping task in the same scene, which is a tougher setting than the static one.

\noindent
\textbf{Metrics.} To evaluate the performance accurately, we follow the grasping metrics from SuctionNet\cite{cao2021suctionnet}.
We adopt $R_{grasp}$, the
ratio of the number of successful grasps to the number total
grasps, and $R_{object}$,  the ratio of the
number of successfully cleared objects to the number of totals. Moreover, if three consecutive grasping attempts fail in a scene, we consider the experiment for that scene terminates.

\begin{table}[h]
\centering
\resizebox{0.8\linewidth}{!}{
\begin{tabular}{cccccc}\toprule
Method        & $R_{grasp}\uparrow$   & $R_{object}\uparrow$  \\ \hline
easy\_handeye\cite{EasyHandEye} & 28/52 = 53.8\%       & 28/32 = 87.5\%             \\
DREAM\cite{Lee2020CameratoRobotPE}         & 26/52 = 50.0\%      & 26/32 = 81.2\%             \\
CenterTrack\cite{CenterTrack}   & 26/60 = 43.3\%       & 26/32 = 81.2\%              \\
CenterNet\cite{CenterNet}     & 29/51 = 56.9\%      & 29/32 = 90.6\%              \\
Ours          & \textbf{32/48 = 66.7\%}       & \textbf{32/32 = 100\%}        
\\ 
\bottomrule
\end{tabular}}
\caption{Quantitative comparison of the performance of different methods applied to robot grasping tasks in the static experiment.}
\label{real_static}
\vspace{-0.9mm}
\end{table}

\begin{table}[h]
\centering
\resizebox{0.8\linewidth}{!}{
\begin{tabular}{cccccc}\toprule
Method        & $R_{grasp}\uparrow$   & $R_{object}\uparrow$ \\\hline
easy\_handeye\cite{EasyHandEye} & -              & -                           \\
DREAM\cite{Lee2020CameratoRobotPE}         & 15/38 = 39.5\% & 15/32 = 46.9\% \\
CenterTrack\cite{CenterTrack}   & 24/51 = 47.1\% & 24/32 = 75.0\% \\
CenterNet\cite{CenterNet}     & 16/42 = 38.0\% & 16/32 = 50.0\%  \\
Ours          & \textbf{30/51 = 58.8\%} & \textbf{30/32 = 93.8\%}   
\\ 
\bottomrule
\end{tabular}}
\caption{Quantitative comparison of different methods' performance applied to robot grasping tasks in the dynamic experiment. The blank in the "easy\_handeye" column is because this traditional calibration method cannot be performed online.}
\label{real_dynamic}
\vspace{-4.4mm}
\end{table}

\noindent
\textbf{Results and Analysis.} 
Table \ref{real_static} and Table \ref{real_dynamic} show the results of applying different camera-to-robot pose estimation methods to the grasping experiments in static and dynamic environments, respectively. In the static experiment, our method achieves a 100\% object grasping success rate, outperforming all other baselines, including traditional hand-eye calibration. In the dynamic experiment, since the camera pose changes after each grasping attempt in the same scene, this places higher demands on the robustness and speed of the camera-to-robot pose estimation. Other methods experience significant drops in grasping success rates, while our method maintains a very high success rate with only a slight decrease compared to the static experiment. These experiments demonstrate the accuracy and stability of our method for camera-to-robot pose estimation.

\section{Conclusion and Future Work}
\label{sec:conclusion}

In this paper, we study the camera-to-robot pose estimation using single-view successive frames from an image sequence. By leveraging the robot structure priors, we use a temporal attention mechanism to efficiently fuse keypoint features from different frames. 
Our method demonstrates significant improvements over synthetic and real-world datasets, strong dominance compared with traditional hand-eye calibration and high accuracy and stability in downstream grasping tasks.
One limitation of our method is that although domain randomisation can narrow the sim-to-real gap to some extent, generalising to arbitrary scenes remains a significant challenge. To address this limitation, future work could explore domain adaptation between real and simulated scenes and fine-tuning in real-world settings.
\section*{Acknowledgements}
\label{sec:acknowledgements}
This project was supported by the National Natural Science Foundation of China (No. 62136001). 
We would also like to thanks the lab mates for the helpful discussion.

{\small
\bibliographystyle{ieee_fullname}
\bibliography{egbib}
}

\end{document}